\documentclass[10pt,twocolumn,letterpaper]{article}

\usepackage[pagenumbers]{cvpr} 

\usepackage{graphicx}
\usepackage{amsmath}
\usepackage{amssymb}
\usepackage{booktabs}
\usepackage{multirow}
\usepackage{makecell}
\usepackage{bm}

\usepackage[pagebackref,breaklinks,colorlinks]{hyperref}

\usepackage[capitalize]{cleveref}
\crefname{section}{Sec.}{Secs.}
\Crefname{section}{Section}{Sections}
\Crefname{table}{Table}{Tables}
\crefname{table}{Tab.}{Tabs.}

\def\cvprPaperID{1621} 
\def\confName{CVPR}
\def\confYear{2022}

\begin{document}

\title{Exploring Dual-task Correlation for Pose Guided Person Image Generation}

\author{Pengze Zhang\textsuperscript{1}, Lingxiao Yang\textsuperscript{1}, Jianhuang Lai\textsuperscript{1,2,3} and Xiaohua Xie\textsuperscript{1,2,3\thanks{Corresponding Author}}\\
\textsuperscript{1}School of Computer Science and Engineering, Sun Yat-Sen University, China\\
\textsuperscript{2}Guangdong Province Key Laboratory of Information Security Technology, China\\
\textsuperscript{3}Key Laboratory of Machine Intelligence and Advanced Computing, Ministry of Education, China\\
{\tt\small zhangpz3@mail2.sysu.edu.cn, \{yanglx9, stsljh, xiexiaoh6\}@mail.sysu.edu.cn}}


\maketitle

\begin{abstract}
Pose Guided Person Image Generation (PGPIG) is the task of transforming a person image from the source pose to a given target pose. Most of the existing methods only focus on the ill-posed source-to-target task and fail to capture reasonable texture mapping. To address this problem, we propose a novel Dual-task Pose Transformer Network (DPTN), which introduces an auxiliary task (i.e., source-to-source task) and exploits the dual-task correlation to promote the performance of PGPIG. The DPTN is of a Siamese structure, containing a source-to-source self-reconstruction branch, and a transformation branch for source-to-target generation. By sharing partial weights between them, the knowledge learned by the source-to-source task can effectively assist the source-to-target learning. Furthermore, we bridge the two branches with a proposed Pose Transformer Module (PTM) to adaptively explore the correlation between features from dual tasks. Such correlation can establish the fine-grained mapping of all the pixels between the sources and the targets, and promote the source texture transmission to enhance the details of the generated target images. Extensive experiments show that our DPTN outperforms state-of-the-arts in terms of both PSNR and LPIPS.  In addition, our DPTN only contains 9.79 million parameters, which is significantly smaller than other approaches. Our code is available at: \url{https://github.com/PangzeCheung/Dual-task-Pose-Transformer-Network}.

\end{abstract}

\section{Introduction}
Pose Guided Person Image Generation (PGPIG) aims to generate person images with arbitrary given poses. It has various applications such as e-commerce, film special effects, person re-identification\cite{9623376,9540780,Wang_Lai_Huang_Xie_2019,9470916,8025424,8765608, 8322217,9507362}, etc. Due to the significant changes in texture and geometry during the pose transfer, PGPIG is still a challenging task.

\begin{figure}[t]
\centering
\includegraphics[width=0.48\textwidth]{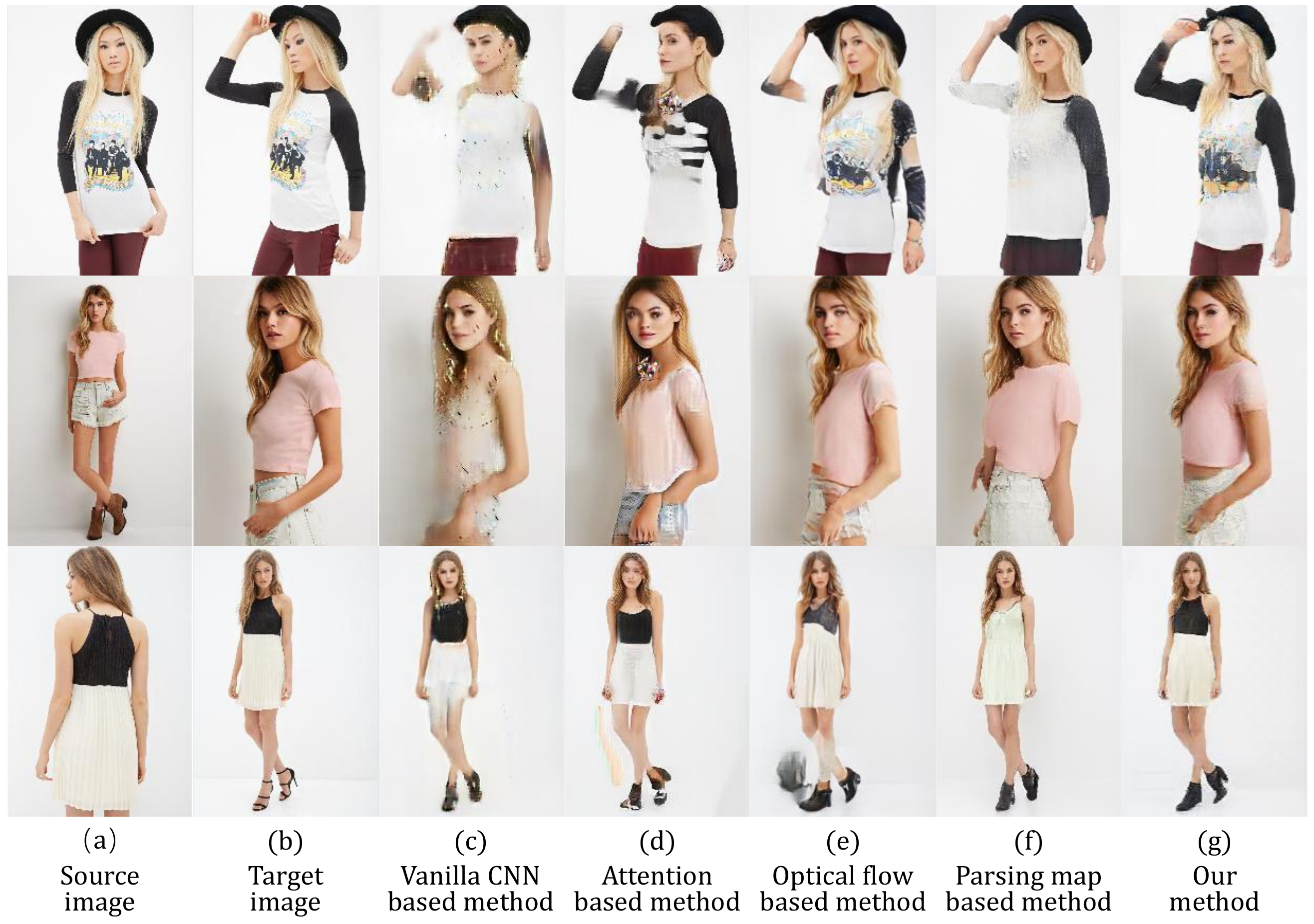} 
\caption{Visual comparison of our method with other approaches, including vanilla CNN based \cite{PG2}, attention based \cite{PATN}, optical flow based \cite{DIST}, and parsing map based \cite{SPIG} method. Compared with other methods, our model can generate more realistic images.}
\label{fig:fig1}
\end{figure}

Driven by the improvement of generative models, e.g., Generative Adversarial Networks (GANs) \cite{NIPS2014_5ca3e9b1} and Variational Autoencoders (VAEs) \cite{VAE}, PGPIG has made great progress. However, early works \cite{PG2,VUnet} are built on vanilla Convolutional Neural Networks (CNNs), which lack the capability to perform complex geometry transformations \cite{NIPS2015_33ceb07b} (see \cref{fig:fig1} (c)). To tackle this problem, attention mechanisms \cite{PATN,XingGAN} and optical flow \cite{DIAF,DIST,tabejamaat2021guided} are applied to improve spatial transformation abilities. Some methods \cite{PISE,SPIG} introduce additional labels such as human parsing maps to provide semantic guidance for pose variations. However, the above mentioned methods solely focus on training the generator $G$ on the \textbf{Source-to-Target Task} that transforms the source image $\bm{x_s}$ from the source pose $\bm{p_s}$ to the target pose $\bm{p_t}$: $G(\bm{x_s}, \bm{p_s}, \bm{p_t}) = \bm{\tilde{x}_t}$. This is an ill-posed problem, making it arduous to train a robust generator. Moreover, the existing methods cannot well capture the reasonable texture mapping between the source and target images, especially when the person undergoes large pose changes. Therefore, those methods often produce unrealistic images, as shown in \cref{fig:fig1} (d)-(f).

In this paper, we seek to utilize an auxiliary task \cite{ruder2017overview} to improve the ill-posed source-to-target transformation. Here, we instantiate the auxiliary task as the \textbf{Source-to-Source Task}, which reconstructs the source image guided by source pose: $G(\bm{x_s}, \bm{p_s}, \bm{p_s}) = \bm{\tilde{x}_s}$. We observe that simultaneously learning the dual tasks (i.e., source-to-target task and source-to-source task) has the following two benefits: (1) Compared with the source-to-target task, the pixel-aligned source-to-source task is easier to learn because it does not require complex spatial transformations. By sharing weights between the dual tasks, the source-to-source task can not only exploit its knowledge to assist the source-to-target task, but also stabilize the training of the whole network. (2) Since the intermediate features in dual tasks are associated with their generated images $\bm{\tilde{x}_s}$ and $\bm{\tilde{x}_t}$ respectively, we can further explore the correlation between the dual tasks to establish the texture transformation from the sources to the targets. In this way, the natural source textures can be readily disseminated to enhance the details of the generated target image.

\begin{figure*}[t]
\centering
\includegraphics[width=1\textwidth]{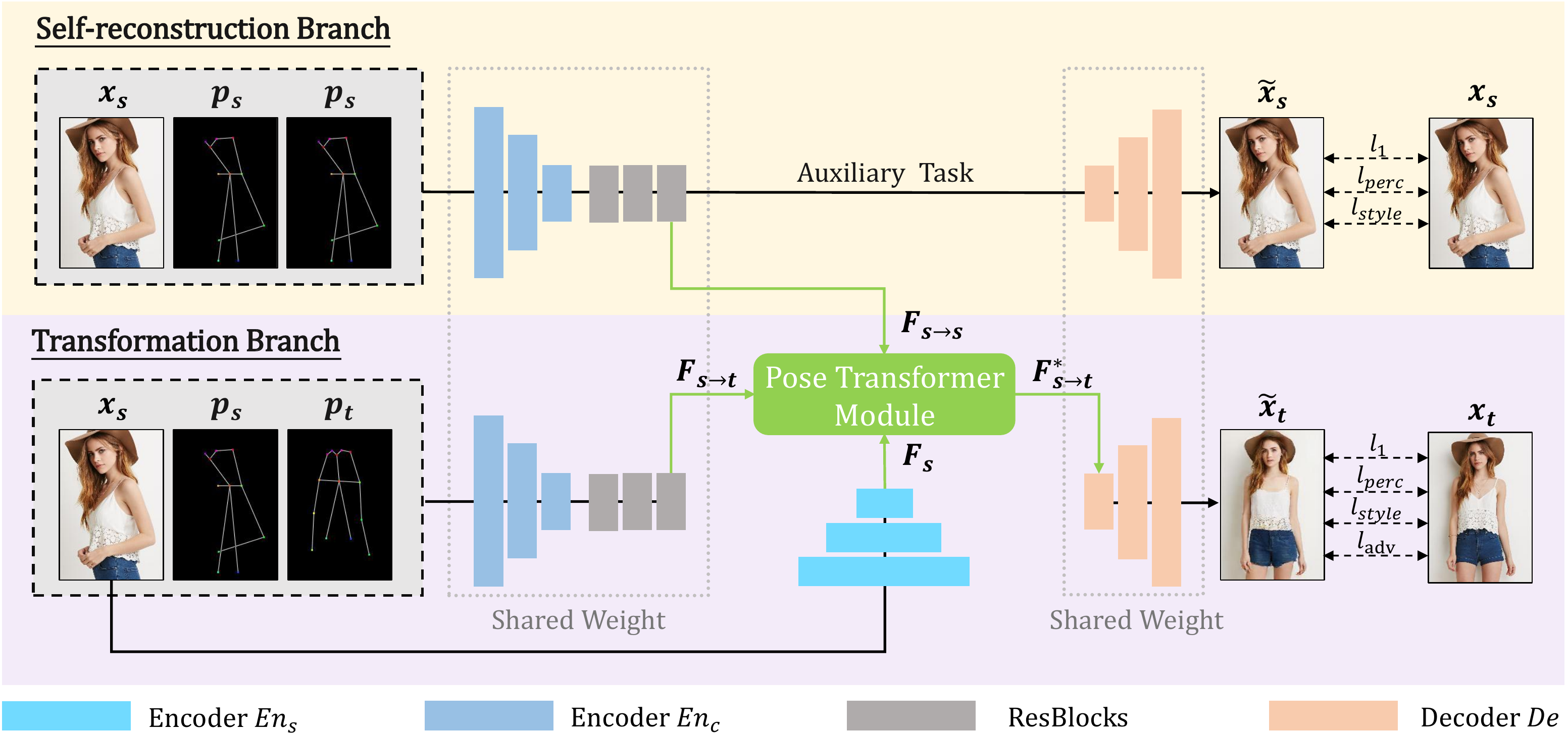} 
\caption{Overview of our model. It contains a self-reconstruction branch for auxiliary source-to-source task, and a transformation branch for source-to-target task. These two branches share partial weights and are communicated by a pose transformer module. }
\label{fig:fig2}
\end{figure*}

Based on these ideas, we propose a novel Dual-task Pose Transformer Network (DPTN) for PGPIG. The architecture of DPTN is shown in \cref{fig:fig2}. Specifically, our DPTN is of a Siamese structure, incorporating two branches: a self-reconstruction branch for the auxiliary source-to-source task and a transformation branch for the source-to-target task. These two branches share partial weights, and are trained simultaneously with different loss functions. By this means, the knowledge learned by the source-to-source task can  directly assist the optimization of the source-to-target task. To explore the correlation between the dual tasks, we bridge the two branches with a novel Pose Transformer Module (PTM). Our PTM consists of several Context Augment Blocks (CABs) and Texture Transfer Blocks (TTBs). CABs first selectively gather the information of the source-to-source task. Then TTBs gradually capture the fine-grained correlation between the features from the dual tasks. With the help of such correlation, TTBs can productively promote the texture transmission from the real source image to the source-to-target task, enabling the synthetic image to preserve more source appearance details. (see \cref{fig:fig1} (g)). In sum, the main contributions are:
\begin{itemize}
\item We propose a novel Dual-task Pose Transformer Network (DPTN), which introduces an auxiliary task (i.e., source-to-source task) by Siamese architecture and exploits its knowledge to improve the PGPIG.
\item We design a Pose Transformer Module (PTM) to explore the dual-task correlation. Such correlation can not only establish the fine-grained mapping between the sources and the targets, but also effectively guide the source texture transmission to further refine the feature in the source-to-target task.
\item Results on the two benchmarks, i.e., DeepFashion \cite{DeepFashion} and Market-1501 \cite{market1501}, have demonstrated that our method exhibits superior performance on PSNR and LPIPS \cite{Zhang_2018_CVPR}. Moreover, our model only contains 9.79 million parameters, which is relatively 91.6\% smaller than the state-of-the-art method SPIG \cite{SPIG}.
\end{itemize}
%

\section{Related Works}

\textbf{Pose guided person image generation.} Ma \emph{et al.} \cite{PG2} generated the fake image in a coarse-to-fine manner. Esser \emph{et al.} \cite{VUnet} combined the VAE and U-net \cite{Unet} to disentangle pose and appearance of the person. However, these methods are based on vanilla CNNs, which cannot handle the complex deformation. To address this problem, Zhu \emph{et al.} \cite{PATN} proposed a Pose Attention Transfer Network (PATN) to optimize the appearance by pose relation. Furthermore, Tang \emph{et al.} \cite{XingGAN} added more crossing ways between the pose and appearance into PATN. Nevertheless, these attention based methods do not explicitly learn the spatial transformation between different poses, losing many source textures. 

To boost the texture transformation,  Li \emph{et al.} \cite{DIAF}, Ren \emph{et al.} \cite{DIST} and Tabejamaat \emph{et al.} \cite{tabejamaat2021guided} proposed to introduce the warping operations to PGPIG. They first estimated the dense optical flow, and then generated images by warping the source image feature. Nevertheless, under the large pose change and occlusion, these methods tended to produce inaccurate optical flow, resulting in unsatisfied images.

Besides, both Zhang \emph{et al.} \cite{PISE} and Lv \emph{et al.} \cite{SPIG} utilized additional human parsing labels to improve the PGPIG. They first predicted the target parsing maps, and then output person images with the help of semantic information. However, the target parsing maps estimated by these methods are often unreliable, which will mislead the generation of the synthetic images. Moreover, pixel-wise annotations are hard to collect, which limits their applications.

In summary, all the above methods only focus on the source-to-target task, and cannot accurately capture the texture mapping between the source and the target images. Contrary to them, we show that introducing the auxiliary source-to-source task through a Siamese structure and simultaneously exploring the dual-task correlation can further improve the performance of PGPIG.

\textbf{Dual-task learning.} Dual-task learning is a popular learning framework for Natural Language Processing (NLP) \cite{NIPS2016_5b69b9cb,DTL,ef88fbee6d014487a9995842fd8a6f64}, which utilizes the different tasks to improve the learning progress. For example, \cite{NIPS2016_5b69b9cb} leveraged the closed-loop of the English-to-French translation and French-to-English translation to enhance each other, making it possible to train translation models without paired data. Different from these methods, our dual tasks refer to the source-to-source task and the source-to-target task. We have verified that the source-to-source learning can promote the training of the source-to-target task in PGPIG.


\textbf{Transformers in vision tasks.} Inspired by the success of transformers \cite{transformer} in NLP, many researchers had applied transformer architecture to computer vision tasks such as image recognition \cite{dosovitskiy2020vit,touvron2020deit}, object detection \cite{detr, deformdetr},  and image generation \cite{jiang2021transgan,hudson2021ganformer}. Specially, for image generation tasks, Jiang \emph{et al.} \cite{jiang2021transgan} built a GAN with a pure transformer-based architecture without convolutions. Hudson \emph{et al.} \cite{hudson2021ganformer} proposed a GANformer to exchange information between image features and latent variables. However, these GANs were designed for unconditional generation tasks, and were not well suitable for conditional generation tasks with complex space deformation (i.e., PGPIG). In this work, inspired by the core idea of the transformer, we design a novel pose transformer module to explore the dual-task correlation. 

\section{Our Approach}
\cref{fig:fig2} shows the overall framework of our DPTN. It mainly contains Siamese branches for the dual tasks and a pose transformer module for exploring the dual-task correlation. In the following sections, we will describe each component of DPTN and loss functions in detail.

\setlength{\tabcolsep}{1.6mm}{
\begin{table}[!t]
\caption{The comparisons of the basic network on whether using the source-to-source learning. Both of the following two results are tested on the source-to-source task. }
\label{tab:taba}
\centering
\begin{tabular}{c|ccc}
\Xhline{1pt}
Learning scheme       & PSNR $\uparrow$           & LPIPS $\downarrow$ \\ \hline
Source-to-target learning                 & 19.1855                    & 0.1962        \\

+ Source-to-source learning                 & 23.7606                    & 0.1468         \\ \Xhline{1pt}
\end{tabular}
\end{table}
}

\subsection{Siamese Structure for Dual Tasks}

\begin{figure*}[!t]
\centering
\includegraphics[width=1\textwidth]{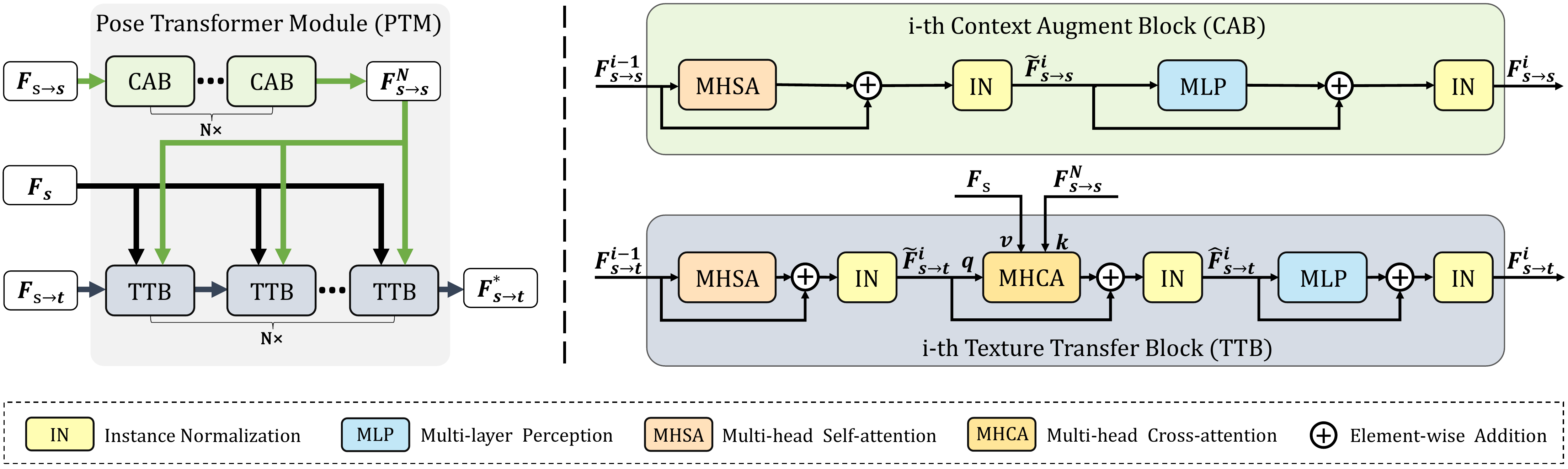} 
\caption{The structure of the Pose Transformer Module (PTM). It contains two types of blocks: Context Augment Block (CAB) and Texture Transfer Block (TTB). The CABs integrate the information of the feature $\bm{F_{s \rightarrow s}}$, while the TTBs transfer the real source image textures from $\bm{F_s}$ to optimize $\bm{F_{s \rightarrow t}}$ by capturing the correlation between features from the dual tasks.}
\label{fig:fig3}
\end{figure*}

Although the existing PGPIG methods attempt to learn the source-to-target transformation through various approaches, we argue that these methods ignore some essential knowledge without the source-to-source learning, thus limiting their potential improvement. To demonstrate this, we conduct an experiment on a basic network (same structure as the self-reconstruction branch in \cref{fig:fig2}, including $En_c$, ResBlocks and $De$) to explore the impact of the source-to-source learning, and show the results tested on the source-to-source task in \cref{tab:taba}. Compared with source-to-target learning, the + source-to-source learning in \cref{tab:taba} only adds the self-reconstruction training, and does not change the basic network structure. It can be seen that there is a significant gap between these two learning schemes. Solely learning from the source-to-target task cannot well reconstruct the source images, and lacks some knowledge of PGPIG. Based on our analysis, in this paper, we add the source-to-source task into PGPIG, and explore its knowledge to assist the source-to-target transformation in the training process.

To achieve this goal, we construct our DPTN with a Siamese architecture, incorporating two branches: a self-reconstruction branch for source-to-source reconstruction, and a transformation branch for source-to-target generation. As shown in \cref{fig:fig2}, the two branches share three parts: an encoder $En_c$, a series of ResBlocks, and a decoder $De$. In more detail, the encoder first extracts the feature of two types of inputs, including the source-to-target input (the concatenation of $\bm{x_s}$, $\bm{p_s}$ and  $\bm{p_t}$) and the source-to-source input (the concatenation of $\bm{x_s}$, $\bm{p_s}$ and $\bm{p_s}$). Then, ResBlocks are applied to gradually perform pose transformation. Outputs of ResBlocks are the feature $\bm{F_{s \rightarrow s}}$ aligned with the source pose, and the transformed feature $\bm{F_{s \rightarrow t}}$ aligned with the target pose. Finally, the $De$ in the self-reconstruction branch accepts the $\bm{F_{s \rightarrow s}}$ to generate fake source image $\bm{\widetilde x_{s}}$, and the $De$ in the transformation branch accepts the refined feature $\bm{F^{*}_{s \rightarrow t}}$ (output of our pose transformer module) to produce the target generated image $\bm{\widetilde x_{t}}$.

In conclusion, the proposed Siamese architecture has the following advantages: (1) Our encoder, ResBlocks and decoder are shared by the dual tasks, so that the learned knowledge can be easily transferred between these tasks. (2) Introducing self-reconstruction branch does not significantly add extra parameters, as most of our model are reused in different tasks. (3) The Siamese architecture enables the intermediate outputs of the dual tasks close in feature distribution, facilitating the PTM in the next section to explore the dual-task correlation.

\subsection{Pose Transformer Module}

In our Siamese structure, we have already obtained feature $\bm{F_{s \rightarrow s}}$ aligned with the source pose $\bm{p_s}$, and $\bm{F_{s \rightarrow t}}$ aligned with the target pose $\bm{p_t}$ respectively. However, since the vanilla CNN based transformation branch (i.e., source-to-target) is hard to handle complex space deformation, $\bm{F_{s \rightarrow t}}$ tends to lose many source appearance details, as shown in \cref{fig:fig7}. To tackle this problem, we propose a novel Pose Transformer Module (PTM), which can further refine $\bm{F_{s \rightarrow t}}$ via capturing the pixel-wise source-to-target correspondence between the features from dual tasks. Our PTM is built upon the Multi-Head Attention (MHA) mechanism. To be self-contained, we briefly introduce MHA as follows: 
\begin{equation}
\label{eq1}
{Attention(\bm{Q}, \bm{K}, \bm{V}) = softmax(\bm{Q}\bm{K}^{T}/{\sqrt{d_k}})\bm{V}},
\end{equation}
\begin{equation}
{\bm{head_i} = Attention(\bm{Q}\bm{W^{i}_q}, \bm{K}\bm{W^{i}_k}, \bm{V}\bm{W^{i}_v})},
\end{equation}
\begin{equation}
\label{eq2}
{MHA(\bm{Q}, \bm{K}, \bm{V}) = concat(\bm{head_1},...,\bm{head_h})}.
\end{equation}
The $\bm{Q}$, $\bm{K}$, $\bm{V}$ are queries, keys and values. ${\bm{W^{i}_q}}$, ${\bm{W^{i}_k}}$, ${\bm{W^{i}_v}}$ are learnable parameters. $h$ is the number of attention heads. $d_k$ is the dimension of the keys. In particular, when $\bm{Q} = \bm{K}$, the MHA functions as Multi-Head Self-Attention (MHSA); otherwise it acts as Multi-Head Cross-Attention (MHCA).

The proposed PTM is shown in \cref{fig:fig3}. Unlike traditional vision transformer \cite{dosovitskiy2020vit}, our PTM adopts a new architecture to explore the relation among triple features (i.e., feature from source-to-source task, source-to-target task and source image texture), making it more suitable for PGPIG. In general, PTM contains two types of blocks: Context Augment Block (CAB) and Texture Transfer Block (TTB), which can be formulated as: ${\bm{F^{N}_{s \rightarrow s}} = CAB(...CAB(\bm{F_{s \rightarrow s}})...)}$, ${\bm{F^{*}_{s \rightarrow t}} = TTB(...TTB(\bm{F_{s \rightarrow t}}, \bm{F^{N}_{s \rightarrow s}}, \bm{F_s})..., \bm{F^{N}_{s \rightarrow s}}, \bm{F_s})}$. The superscript denotes the index of the feature. $N$ is the number of the blocks. $\bm{F_{s \rightarrow s}} \! = \!\bm{F^{0}_{s \rightarrow s}}$, $\bm{F_{s \rightarrow t}}\! = \!\bm{F^{0}_{s \rightarrow t}}$, and $\bm{F^{*}_{s \rightarrow t}}\! = \!\bm{F^{N}_{s \rightarrow t}}$. $\bm{F_s}$ is the source image feature obtained by an additional encoder $En_s$. In the proposed PTM, the CABs gradually integrate the information of the feature $\bm{F_{s \rightarrow s}}$ from the self-reconstruction branch and produce $\bm{F^{N}_{s \rightarrow s}}$. Then, each TTB combines three kinds of features: the source image texture feature $\bm{F_s}$, the integrated source-to-source feature $\bm{F_{s \rightarrow s}}$, and the previous TTB output $\bm{F_{s \rightarrow t}}$. This combination is achieved by an MHCA module to capture the correlation among all inputs. Next, we will present the structure of the CAB and TTB respectively.

\subsubsection{Context Augment Block}
The structure of the $i$-th CAB is shown in the right-top of \cref{fig:fig3}. It first applies an MHSA unit with residual connection to adaptively enhance the contextual representation of the input feature ${\bm{F^{i-1}_{s \rightarrow s}}}$:
\begin{equation}
\label{eq3}
{\bm{\tilde{F}^{i}_{s \rightarrow s}} = IN(\bm{F^{i-1}_{s \rightarrow s}} + MHSA(\bm{F^{i-1}_{s \rightarrow s}}, \bm{F^{i-1}_{s \rightarrow s}}, \bm{F^{i-1}_{s \rightarrow s}}))},
\end{equation}
where IN is the instance normalization \cite{ulyanov2017improved}. Then a Multi-Layer Perceptron (MLP) module with multiple fully connected layers is used to increase the capacity in CAB:
\begin{equation}
\label{eq4}
{\bm{F^{i}_{s \rightarrow s}} = IN(\bm{\tilde{F}^{i}_{s \rightarrow s}} + MLP(\bm{\tilde{F}^{i}_{s \rightarrow s}}))}.
\end{equation}
After $N$ CABs, we obtain the final refined feature $\bm{F^{N}_{s \rightarrow s}}$ and add this feature into each TTB for source-to-target task.

\subsubsection{Texture Transfer Block}
The structure of the $i$-th TTB is shown in the right-bottom of \cref{fig:fig3}. First, MHSA is applied to selectively focus on the key information of the transformation branch feature $\bm{F^{i-1}_{s \rightarrow t}}$:
\begin{equation}
\label{eq5}
{\bm{\tilde{F}^{i}_{s \rightarrow t}} = IN(\bm{F^{i-1}_{s \rightarrow t}} + MHSA(\bm{F^{i-1}_{s \rightarrow t}}, \bm{F^{i-1}_{s \rightarrow t}}, \bm{F^{i-1}_{s \rightarrow t}}))}.
\end{equation}
Then an MHCA unit is employed to build correlation of $\bm{\tilde{F}^{i}_{s \rightarrow t}}$, $\bm{F^{N}_{s \rightarrow s}}$, and $\bm{{F}_{s}}$. Specifically, we employ $\bm{\tilde{F}^{i}_{s \rightarrow t}}$ as queries and $\bm{F^{N}_{s \rightarrow s}}$ as keys to calculate the pixel-wise similarity between the sources and the targets. With the aid of such similarity, $\bm{F_{s}}$ is used as values in MHCA to transmit the real source textures to refine $\bm{\tilde{F}^{i}_{s \rightarrow t}}$. This procedure can be written as:
\begin{equation}
\label{eq6}
{\bm{\hat{F}^{i}_{s \rightarrow t}} = IN(\bm{\tilde{F}^{i}_{s \rightarrow t}} + MHCA(\bm{\tilde{F}^{i}_{s \rightarrow t}}, \bm{F^{N}_{s \rightarrow s}}, \bm{F_{s}}))}.
\end{equation}
In this way, $\bm{\hat{F}^{i}_{s \rightarrow t}}$ carries more real source textures, which will foster the transformation branch to generate more delicate patterns. Finally, similar to CAB, the $i$-th TTB output $\bm{F^{i}_{s \rightarrow t}}$ is obtained as follows:
\begin{equation}
\label{eq7}
{\bm{F^{i}_{s \rightarrow t}} = IN(\bm{\hat{F}^{i}_{s \rightarrow t}} + MLP(\bm{\hat{F}^{i}_{s \rightarrow t}}))}.
\end{equation}
After $N$ time TTB blocks, the final output feature $\bm{F^{N}_{s \rightarrow t}}$ will be fed into the decoder $De$ to generate target image $\bm{\tilde{x}_t}$.

\subsection{Loss Functions}
Our network contains two branches for the source-to-source task and source-to-target task. Thus, the overall loss function can be simply formulated as:
\begin{equation}
\label{eq8}
{\mathcal{L} = \mathcal{L}_{s \rightarrow s} + \mathcal{L}_{s \rightarrow t}},
\end{equation}
where $\mathcal{L}_{s \rightarrow s}$ and $\mathcal{L}_{s \rightarrow t}$ stand for the loss of the dual tasks respectively. Both of them contain an ${l_1}$ loss ${\mathcal{L}_{l_1}}$, a perceptual loss ${\mathcal{L}_{perc}}$ and a style loss ${\mathcal{L}_{style}}$. In addition, we apply an additional adversarial loss  ${\mathcal{L}_{adv}}$ in the source-to-target task to produce more realistic textures. In sum, $\mathcal{L}_{s \rightarrow s}$ and $\mathcal{L}_{s \rightarrow t}$ can be written as:
\begin{equation}
\label{eq:eq9}
{\mathcal{L}_{s \rightarrow s} = \mathcal{\lambda}_{l_1}\mathcal{L}^{s}_{l_1} + \mathcal{\lambda}_{perc}\mathcal{L}^{s}_{perc} + \mathcal{\lambda}_{style}\mathcal{L}^{s}_{style}},
\end{equation}
\begin{equation}
\label{eq:eq10}
{\mathcal{L}_{s \rightarrow t} = \lambda_{l_1}\mathcal{L}^{t}_{l_1} + \lambda_{perc}\mathcal{L}^{t}_{perc} + \lambda_{style}\mathcal{L}^{t}_{style} + \lambda_{adv}\mathcal{L}_{adv}},
\end{equation}
where $\lambda_{l_1}$, $\lambda_{perc}$, $\lambda_{style}$ and $\lambda_{adv}$ are the loss weights for the dual tasks. Specifically, the ${l_1}$ loss penalizes the ${l_1}$ distance between the generated image and the ground truth:
\begin{equation}
\label{eq11}
{\mathcal{L}^{d}_{l_1} = {\Vert{\bm{x_d} - \bm{\tilde{x}_d}}\Vert}_1},
\end{equation}
where $d \in \{s,t\}$ represents the source or the target data. The perceptual loss \cite{feifei} calculates the feature distance:
\begin{equation}
\label{eq12}
{\mathcal{L}^{d}_{perc} = \sum\limits_{i}{{\Vert{\phi_{i}(\bm{x_d}) -  \phi_{i}(\bm{\tilde{x}_d}})\Vert}_1}},
\end{equation}
where $\phi_{i}$ denotes the $i$-th feature from VGG network \cite{VGG}. The style loss \cite{feifei} compares the style similarity between images:
\begin{equation}
\label{eq13}
{\mathcal{L}^{d}_{style} = \sum\limits_{j}{{\Vert{\mathop{Gram}\nolimits_{j}^{\phi}(\bm{x_d}) -  \mathop{Gram}\nolimits_{j}^{\phi}(\bm{\tilde{x}_d}})\Vert}_1}},
\end{equation}
where $\mathop{Gram}^{\phi}_{j}$ is the Gram matrix of feature $\phi_j$. Finally, the adversarial loss with a discriminator $D$ is employed to penalize the distribution difference between the generated target image ${\bm{\tilde{x}_t}}$ and the ground truth ${\bm{{x}_t}}$:
\begin{equation}
\label{eq14}
{\mathcal{L}_{adv} = \mathbb{E}[\log (1-D(\bm{\tilde{x}_t}))] + \mathbb{E}[\log D(\bm{x_t})]}.
\end{equation}

\setlength{\tabcolsep}{1.3mm}{
\begin{table*}[t]
\caption{Quantitative comparisons of image quality and model size with several state-of-the-art methods. * denotes the method using additional human parsing labels. The best and second best results are shown in bold and underline respectively.}
\begin{tabular}{c|cccc|cccc|c}
\Xhline{1pt}
{\multirow{2}{*}{Model}} & \multicolumn{4}{c|}{DeepFashion}    & \multicolumn{4}{c|}{Market1501}      & \multirow{2}{*}{\begin{tabular}[c]{@{}c@{}}Number of\\ Parameters $\downarrow$\end{tabular}} \\ \Xcline{2-9}{0.5pt}
                      & SSIM $\uparrow$   & PSNR $\uparrow$    & FID $\downarrow$     & LPIPS $\downarrow$ & SSIM $\uparrow$   & PSNR $\uparrow$    & FID $\downarrow$    & LPIPS $\downarrow$ &                                                                                 \\ \hline
PG2 \cite{PG2} (NeurIPS'17)          & 0.7730 & 17.5324 & 49.5674 & 0.2928 & 0.2704 & 14.1749 & 86.0288  & 0.3619 & 437.09 M                                                                        \\
VU-net \cite{VUnet} (CVPR'18)        & 0.7639 & 17.6582 & 15.5747 & 0.2415 & 0.2665 & 14.4220 & 44.2743  & 0.3285 & 139.36 M                                                                        \\
DSC\cite{DSC} (CVPR'18)              & 0.7682 & 18.0990 & 21.2686 & 0.2440 & \underline{0.3054} & 14.3081 & 27.0118  & 0.3029 & 82.08 M                                                                         \\
PATN \cite{PATN} (CVPR'19)      & 0.7717 & 18.2543 & 20.7500 & 0.2536 & 0.2818 & 14.2622 & 22.6814  & 0.3194 & 41.36 M                                                                         \\
DIAF \cite{DIAF} (CVPR'19)          & 0.7738 & 16.9004 & 14.8825 & 0.2388 & 0.3052 & 14.2011 & 32.8787  & 0.3059 & 49.58 M                                                                         \\
DIST \cite{DIST} (CVPR'20)             & 0.7677 & 18.5737 & \textbf{10.8429} & 0.2258 & 0.2808 & 14.3368 & \underline{19.7403}  & 0.2815 & \underline{14.04 M}                                                                         \\
XingGAN \cite{XingGAN} (ECCV'20)             & 0.7706 & 17.9226 & 39.3194 & 0.2928 & 0.3044 & 14.4458 & 22.5198  & 0.3058 & 42.77 M                                                                         \\
PISE$^{*}$ \cite{PISE} (CVPR'21)            & 0.7682 & 18.5208 & 11.5144 & \underline{0.2080} &     —    &   —       &    —       &    —     & 64.01 M                                                                         \\
SPIG$^{*}$ \cite{SPIG} (CVPR'21)             & \underline{0.7758} & \underline{18.5867} & 12.7027 & 0.2102 & \textbf{0.3139} & \underline{14.4894} & 23.0573  & \underline{0.2777} & 117.13 M                                                                        \\ \hline
\textbf{Ours}                   & \textbf{0.7782} & \textbf{19.1492} & \underline{11.4664} & \textbf{0.1957} & 0.2854 & \textbf{14.5207} & \textbf{18.9946} & \textbf{0.2711} & \textbf{9.79} M                                                                          \\ \Xhline{1pt}
\end{tabular}
\label{tab:tab1}
\end{table*}
}

\section{Experiments}
\subsection{Implementation Details}
We evaluate our proposed model on two datasets: DeepFashion \cite{DeepFashion} and Market1501 \cite{market1501}. The DeepFashion dataset contains 52,712 high quality  in-shop clothes images (256 $\times$ 176) with clean backgrounds, while the Market1501 dataset contains 32,668 low-resolution images (128 $\times$ 64) with various illumination and viewpoints. For a fair comparison, we split the datasets with the same setting as \cite{PATN}. It collects 101,966 training pairs and 8,570 testing pairs for DeepFashion, and 263,632 training pairs and 12,000 testing pairs for Market1501. In addition, the human pose keypoints are extracted from Human Pose Estimator (HPE) \cite{openpose}.

In our experiment, Adam optimizer \cite{adam} is adopted to train the proposed DPTN with the learning rate 1e-4. We choose $h  = 2$ and $N=2$ in the PTM on both datasets. For the loss functions in \cref{eq:eq9} and \cref{eq:eq10}, we set $\mathcal{\lambda}_{l_1} = 2.5$, $\mathcal{\lambda}_{perc} = 0.25$, $\mathcal{\lambda}_{style} = 250$, $\mathcal{\lambda}_{adv} = 2$.

\subsection{Metrics}
Following previous works \cite{PATN,SPIG}, we adopt Structural Similarity Index Measure (SSIM) \cite{SSIM}, Peak Signal-to-Noise Ratio (PSNR), Fréchet Inception Distance (FID) \cite{FID} and Learned Perceptual Image Patch Similarity (LPIPS) \cite{Zhang_2018_CVPR} as evaluation metrics. Moreover, we use rank-k and Mean Average Precision (MAP) to further test the texture consistency between the source images and the generated target images through the state-of-the-art re-identification (re-id) platform FastReID \cite{he2020fastreid}. More precisely, we train a re-id model on the training set. Then we use the generated images as the query set and the real images as the gallery set to calculate the metrics. High rank-k and MAP indicate that the generated images do not lose much source appearance, and can be easily recognized by the current re-id system.

\subsection{Comparison with Previous Work}

\subsubsection{Quantitative Comparison} 
We compare our method with several state-of-the-art methods, including PG2 \cite{PG2}, VU-net \cite{VUnet}, DSC\cite{DSC}, PATN \cite{PATN}, DIAF \cite{DIAF}, DIST \cite{DIST}, XingGAN \cite{XingGAN}, PISE \cite{PISE} and SPIG \cite{SPIG}. \cref{tab:tab1} shows the quantitative results on image quality and model size. As one can see, our method achieves seven best and one second-best results among all compared methods, including PISE and SPIG using additional parsing labels. This verifies the superiority of our DPTN in generating high quality images. In addition, our DPTN only contains 9.79 M parameters, which is 91.6\% lower than that of SPIG (117.13 M). It clearly demonstrates the efficiency of our method in modeling pose transformations.

\cref{tab:tab2} provides the comparison of the texture consistency on DeepFashion. First, we train the re-id system on the DeepFashion training set. As shown in the last row of \cref{tab:tab2}, this re-id system achieves 99.08\% rank-1 score. Then, the same re-id system is applied to identify the person in the fake images generated by different methods. From the result, we can find that our method surpasses others in all four metrics. In particular, we promote the best rank-1 performance of previous works by 3\%. This indicates that the images generated by our DPTN can effectively maintain the discriminative texture of the source person.

\begin{figure*}[t]
\centering
\includegraphics[width=1\textwidth]{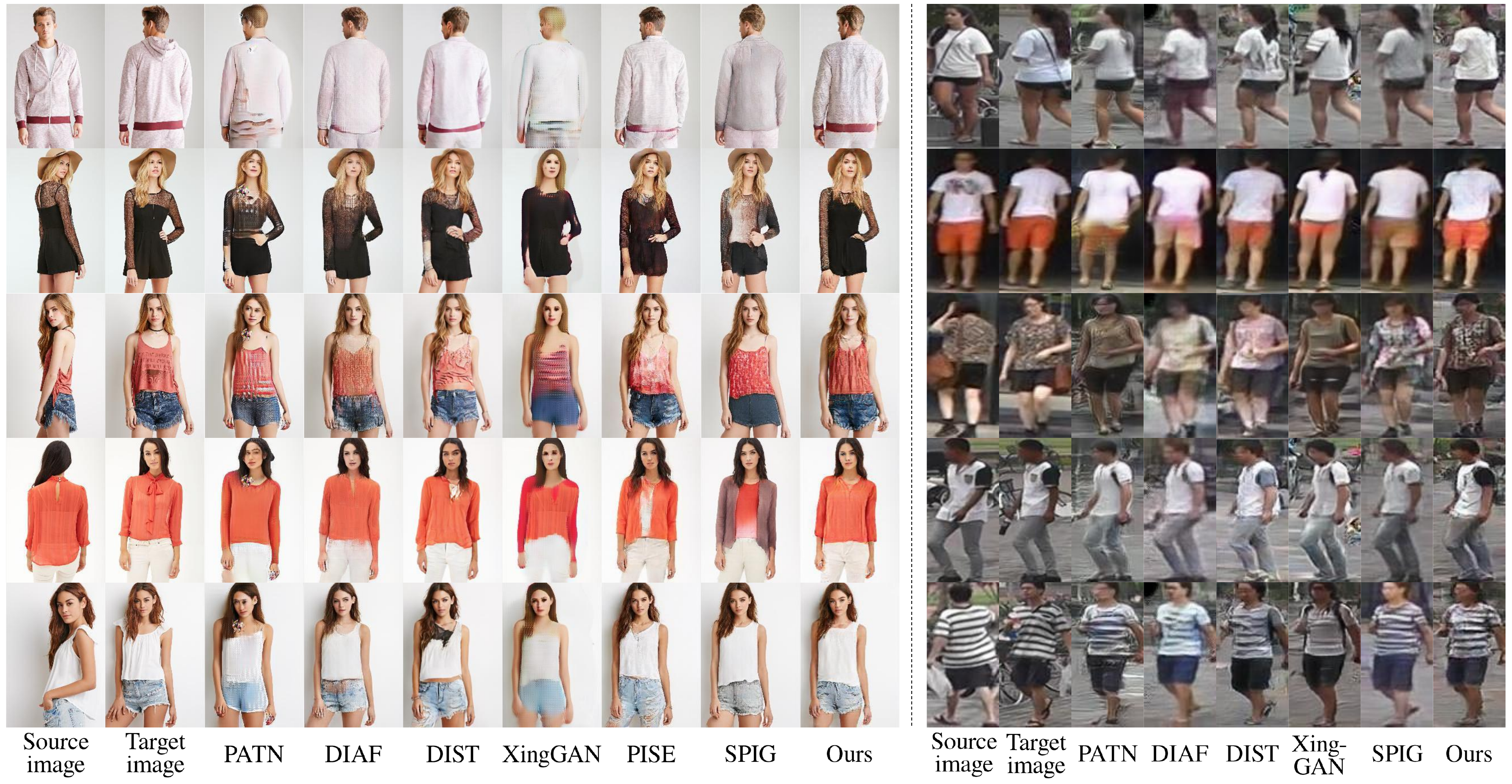} 
\caption{Qualitative comparison with several state-of-the-art methods on DeepFashion (Left) and Market1501 (Right).}
\label{fig:fig4}
\end{figure*}

\setlength{\tabcolsep}{2.0mm}{
\begin{table}[!t]
\caption{Quantitative comparisons of texture consistency.  The best results are shown in bold.}
\centering
\begin{tabular}{c|cccc}
\Xhline{1pt}
Methods       & rank-1 $\uparrow$           & rank-5 $\uparrow$          & rank-10 $\uparrow$             & MAP $\uparrow$ \\ \hline
PG2       & 60.12\%          & 75.44\%          & 81.95\%          & 59.20\%        \\
VU-net       & 73.49\%          & 87.49\%          & 91.97\%          & 72.01\%        \\
DSC           & 94.17\%          & 98.19\%          & 99.08\%          & 90.40\%         \\
PATN          & 74.35\%          & 87.95\%          & 92.37\%          & 73.17\%         \\
DIAF          & 94.87\%          & 98.02\%          & 99.13\%          & 91.45\%         \\ 
DIST          & 90.84\%          & 96.64\%          & 98.11\%          & 87.56\%         \\ 
XingGAN          & 59.63\%          & 72.48\%          & 81.19\%          & 58.36\%         \\ 
PISE          & 90.09\%          & 96.35\%          & 98.02\%          & 87.22\%         \\
SPIG          & 94.43\%          & 98.23\%          & 99.04\%          & 91.60\%         \\ \hline

\textbf{Ours}         & \textbf{97.69\%}          & \textbf{99.35\%}          & \textbf{99.63\%}          & \textbf{95.04\%}         \\ \hline
Real Data & 99.08\% & 99.80\% & 99.88\% & 98.40\% \\ \Xhline{1pt}
\end{tabular}
\label{tab:tab2}
\end{table}
}

\subsubsection{Qualitative Comparison}
The qualitative comparison results are shown in \cref{fig:fig4}. For the DeepFashion dataset, the attention based methods PATN and XingGAN tend to generate blurred images (see 1st and 2nd rows). DIAF and DIST attempt to promote the texture transfer by using optical flow. However, in the case of large pose changes, their predicted optical flow fails to represent such complex deformation, resulting in unacceptable results (see 2nd and 3rd rows). PISE and SPIG introduce the additional semantic parsing map to ease the difficulty of PGPIG. Nevertheless, the target parsing maps estimated by these methods are often inaccurate, which will mislead the generation of the synthetic images. For example, in the 4th row on the left of \cref{fig:fig4}, PISE and SPIG improperly generate jackets in the synthetic images. Unlike the aforementioned methods, our DPTN optimizes the source-to-target task with the help of the auxiliary source-to-source task, making the generated image more vibrant. On the Market1501 dataset, our DPTN can still generate finer and more vivid textures than other methods. For instance, in the 4th row on the right of \cref{fig:fig4}, only our method retains the garment pattern of the source image.

\subsection{Ablation Study}
We conduct a series of experiments on DeepFashion to verify the contribution of each component in our model. The various options for removing the corresponding components from our full model are listed as follows.

\noindent
\textbf{The model without Dual-Task Learning (w/o DTL).} This model is similar to the existing methods that only focus on the source-to-target task. The entire self-reconstruction branch, including $De$ and the loss function, is removed.

\begin{figure}[!t]
\centering
\includegraphics[width=0.48\textwidth]{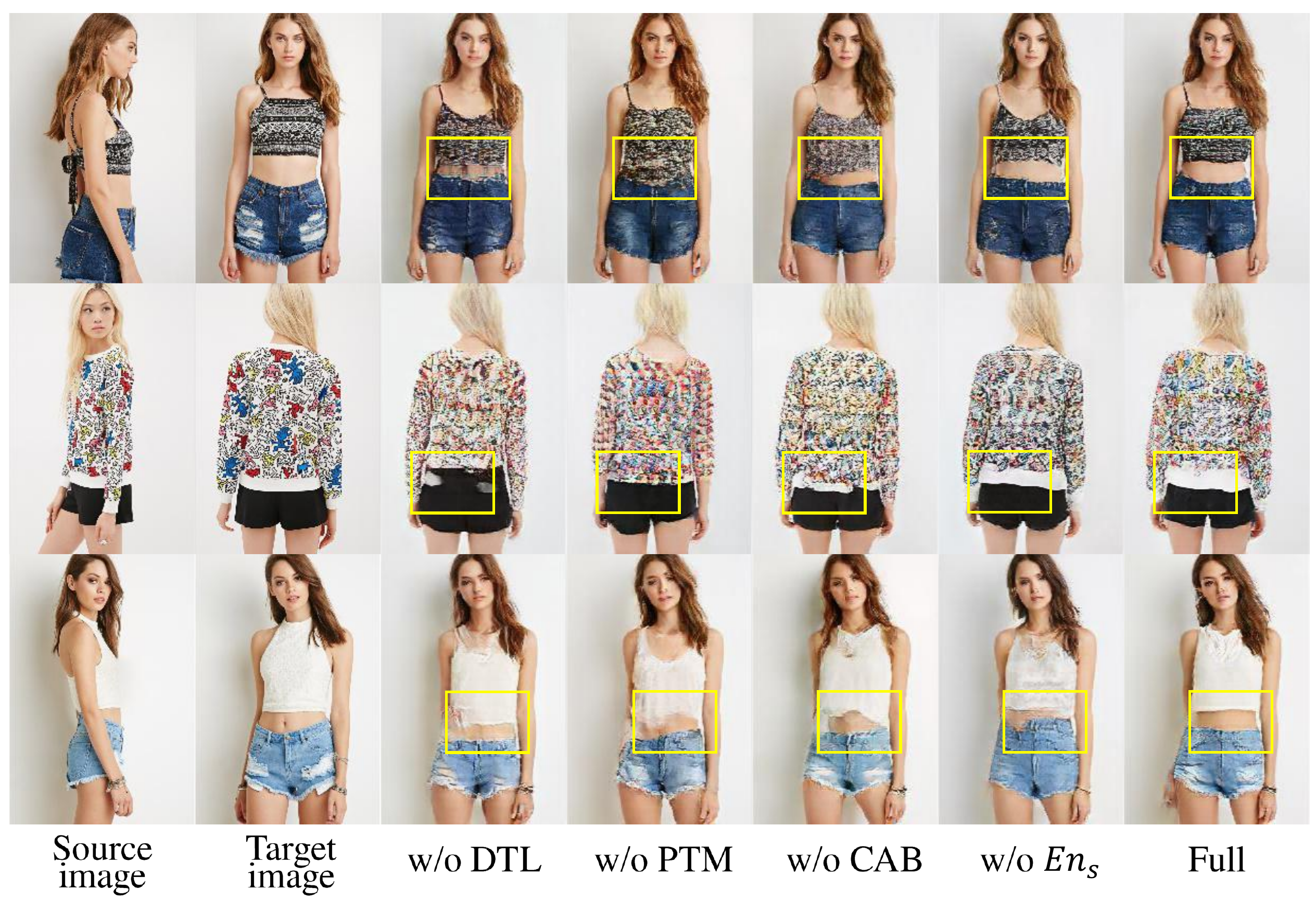} 
\caption{Qualitative comparison of the ablation study.}
\label{fig:fig5}
\end{figure}

\noindent
\textbf{The model without Pose Transformer Module (w/o PTM).} This model removes the PTM. In this way, the source-to-target branch will lack the guidance of the dual-task correlation, and will directly produce the target generated image ($\bm{\widetilde x_{t}}$) from $\bm{F_{s \rightarrow t}}$.

\noindent
\textbf{The model without Contextual Augment Blocks (w/o CABs).}
This model removes CABs in the PTM. In this way, the feature from the source-to-source task ($\bm{F_{s\rightarrow s}}$) will be simply fed into TTBs to calculate the dual-task correlation.

\noindent
\textbf{The model without encoder $\bm{En_s}$ (w/o $\bm{En_s}$).}
This model removes the encoder $En_s$, and directly uses the feature $\bm{F_{s\rightarrow s}}$ as the value in MHCA.

\noindent
\textbf{Full Model (Full).}
We use our proposed dual-task pose transformer network in this model.

\setlength{\tabcolsep}{2.0mm}{
\begin{table}[!t]
\renewcommand{\arraystretch}{1.05}
\caption{Quantitative comparisons of ablation study on the DeepFashion dataset. The best results are shown in bold.}
\label{tab:tab3}
\centering
\begin{tabular}{c|cccc}
\Xhline{1pt}
Methods       & SSIM $\uparrow$           & PSNR $\uparrow$          & FID $\downarrow$             & LPIPS $\downarrow$ \\ \hline
w/o DTL       & 0.7713          & 18.8134          & 14.7168          & 0.2143        \\
w/o PTM       & 0.7755          & 18.8503          & 15.5281           & 0.2195        \\
w/o CABs       & 0.7760          & 19.0489          & 12.0932          & 0.1989        \\
w/o $En_s$           & 0.7778          & 19.1084          & 12.6858          & 0.1976         \\ 
 \hline

\textbf{Full}         & \textbf{0.7782}          & \textbf{19.1492}          & \textbf{11.4664}          & \textbf{0.1957}         \\ \Xhline{1pt}
\end{tabular}
\end{table}
}

\begin{figure}[!t]
\centering
\includegraphics[width=0.48\textwidth]{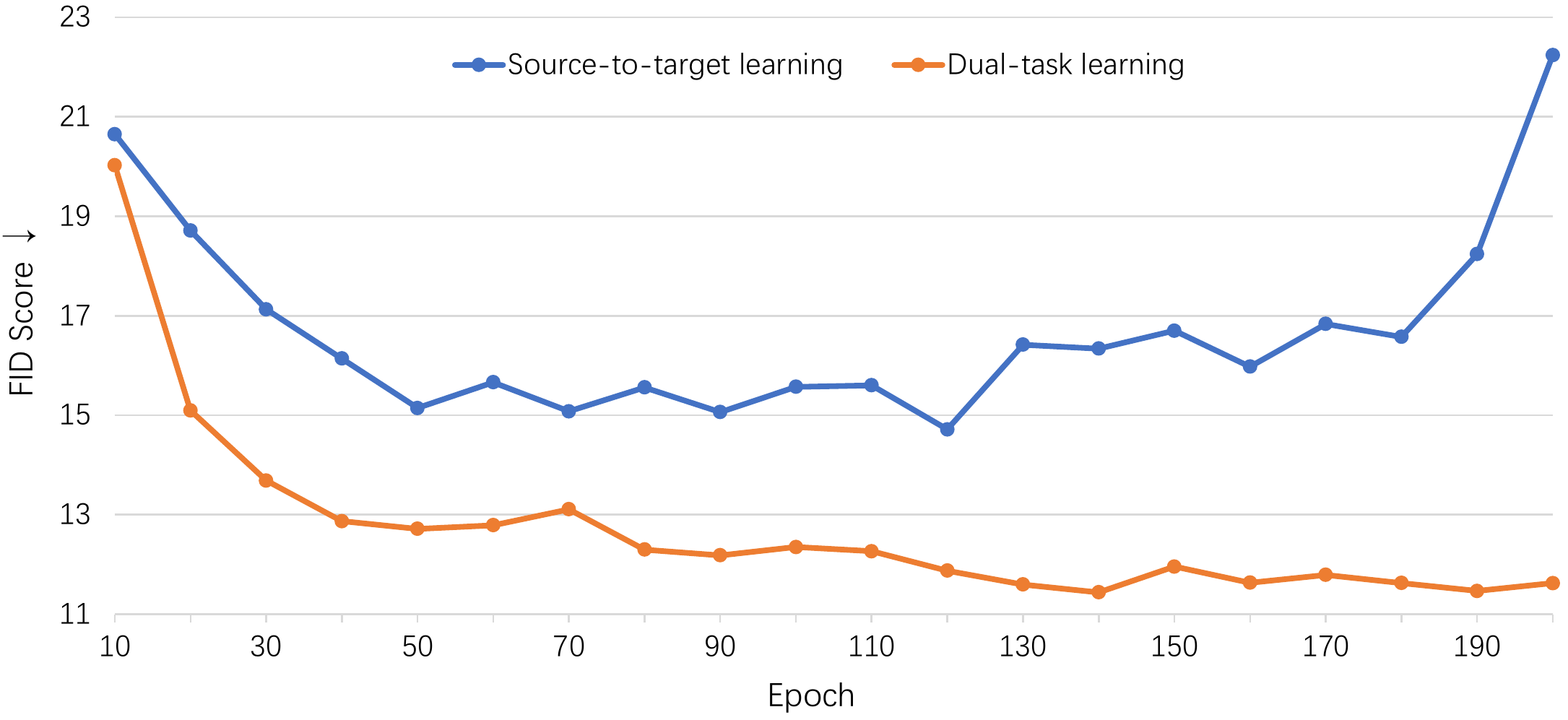} 
\caption{Learning curves of FID score by using source-to-target learning and dual-task learning on DeepFashion dataset.}
\label{fig:fig6}
\end{figure}

\cref{fig:fig5} and \cref{tab:tab3} show the qualitative and quantitative results of the ablation study. As shown in \cref{fig:fig5}, we can see that (1) Compared with the full model, the model w/o DTL is unstable and tends to generate heavy artifacts. This demonstrates the significance of the source-to-source task during the training process. (2) Lack of texture mapping between the sources and the targets, the model w/o PTM cannot well utilize the textures of the real source image, resulting in blurred images. (3) For the model w/o CABs, the information of the source-to-source task is not well integrated, which misleads the source-to-target task to generate unrealistic patterns. (4) The images generated by the model w/o $En_s$ lose many appearance details, verifying the effect of $En_s$ in supplying fine source textures for PTM. (5) Compared with others, our full model can not only generate satisfactory global appearance but also produce realistic local textures. In addition, the quantitative comparison in \cref{tab:tab3} further demonstrates the effectiveness of our full model.

\subsection{Effect of dual-task learning on training stability}
To explore the influence of dual-task learning on training stability, following \cite{miyato2018spectral}, we visualize the learning curves of FID score under the source-to-target learning and dual-task learning in \cref{fig:fig6}. We can see that the FID score of the DPTN with solely source-to-target learning plateaus around 50 epochs, while the DPTN with dual-task learning continues to improve even afterward. This verifies that the knowledge brought by the source-to-source learning can effectively assist the learning of the source-to-target task. In addition, compared with dual-task learning, the DPTN with solely source-to-target learning tends to collapse after 130 epochs. This shows that by sharing partial weights between the dual tasks, the training of the easier source-to-source task can stabilize the training of the whole network to a certain extent, so as to better optimize the source-to-target task.

\subsection{Visualization of PTM}
\begin{figure}[!t]
\centering
\includegraphics[width=0.48\textwidth]{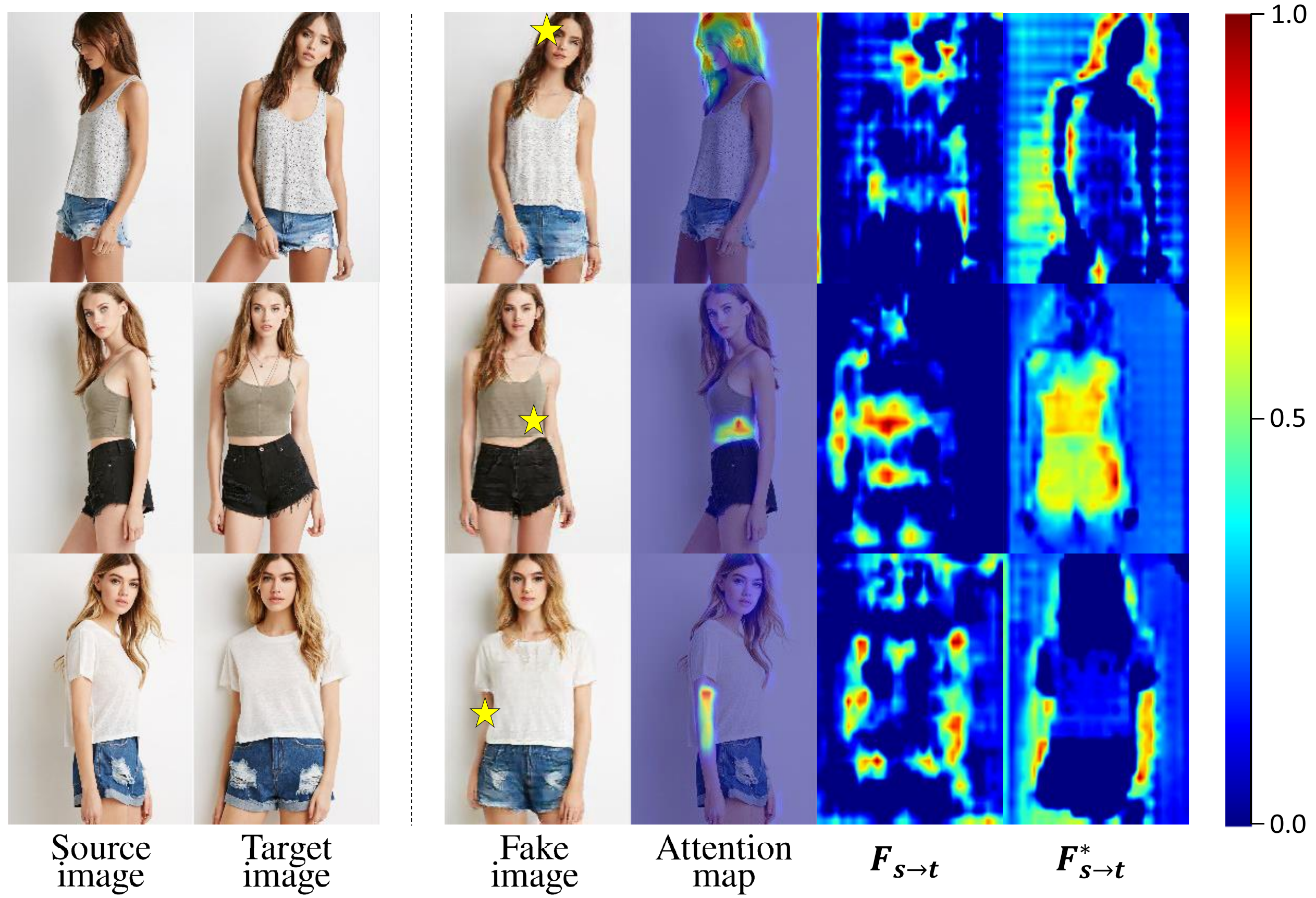} 
\caption{Visualization of the attention weight in MHCA and the heatmaps of $\bm{F_{s\rightarrow t}}$ and $\bm{F^{*}_{s\rightarrow t}}$. The yellow star on the fake image represents the query position.}
\label{fig:fig7}
\end{figure}

To explore how the PTM works in our framework, we also visualize the attention weight in MHCA as well as the heatmaps of the $\bm{F_{s\rightarrow t}}$ and $\bm{F^{*}_{s\rightarrow t}}$ in \cref{fig:fig7}. As one can see, the attention weight obtained in the PTM can accurately focus the area related to the query position. This verifies that our PTM can effectively explore the pixel-wise transformation between the sources and targets. In addition, compared with the heatmap of the $\bm{F_{s\rightarrow t}}$, the $\bm{F^{*}_{s\rightarrow t}}$ produced by PTM contains more appearance cues. This manifests that our PTM can transfer the natural source textures to refine $\bm{F_{s\rightarrow t}}$, and facilitate the source-to-target task to generate more realistic details.

\section{Conclusions}
In this paper, we propose a novel Dual-task Pose Transformer Network (DPTN) for PGPIG. Unlike most of the existing methods only focusing on the source-to-target task, our DPTN introduces an auxiliary task (i.e., source-to-source task) by a Siamese architecture, and exploits its knowledge to assist the source-to-target learning. Moreover, we carefully design a Pose Transformer Model (PTM) to explore the correlation between the dual tasks. Such correlation can be employed as a strong guidance for transferring source textures to the target generated image. Both the quantitative and qualitative results show that the proposed DPTN can improve upon prior PGPIG methods. 

\paragraph{Acknowledgments.} This project is supported by the Key-Area Research and Development Program of Guangdong Province (2019B010155003), and the National Natural Science Foundation of China (62072482).

{\small
\bibliographystyle{ieee_fullname}
\bibliography{egbib}

\begin{thebibliography}{10}\itemsep=-1pt

\bibitem{openpose}
Zhe Cao, Tomas Simon, Shih-En Wei, and Yaser Sheikh.
\newblock Realtime multi-person 2d pose estimation using part affinity fields.
\newblock In {\em Proceedings of the IEEE Conference on computer Vision and
  Pattern Recognition (CVPR)}, pages 1302--1310, July 2017.

\bibitem{detr}
Nicolas Carion, Francisco Massa, Gabriel Synnaeve, Nicolas Usunier, Alexander
  Kirillov, and Sergey Zagoruyko.
\newblock End-to-end object detection with transformers.
\newblock In {\em European Conference on Computer Vision (ECCV)}, pages
  213--229, 2020.

\bibitem{dosovitskiy2020vit}
Alexey Dosovitskiy, Lucas Beyer, Alexander Kolesnikov, Dirk Weissenborn,
  Xiaohua Zhai, Thomas Unterthiner, Mostafa Dehghani, Matthias Minderer, Georg
  Heigold, Sylvain Gelly, Jakob Uszkoreit, and Neil Houlsby.
\newblock An image is worth 16x16 words: Transformers for image recognition at
  scale.
\newblock In {\em International Conference on Learning Representations (ICLR)},
  2021.

\bibitem{VUnet}
Patrick Esser, Ekaterina Sutter, and Björn Ommer.
\newblock A variational u-net for conditional appearance and shape generation.
\newblock In {\em Proceedings of the IEEE Conference on computer Vision and
  Pattern Recognition (CVPR)}, pages 8857--8866, June 2018.

\bibitem{8322217}
Zhanxiang Feng, Jianhuang Lai, and Xiaohua Xie.
\newblock Learning view-specific deep networks for person re-identification.
\newblock {\em IEEE Transactions on Image Processing}, 27(7):3472--3483, 2018.

\bibitem{8765608}
Zhanxiang Feng, Jianhuang Lai, and Xiaohua Xie.
\newblock Learning modality-specific representations for visible-infrared
  person re-identification.
\newblock {\em IEEE Transactions on Image Processing}, 29:579--590, 2020.

\bibitem{9507362}
Zhanxiang Feng, Jianhuang Lai, and Xiaohua Xie.
\newblock Resolution-aware knowledge distillation for efficient inference.
\newblock {\em IEEE Transactions on Image Processing}, 30:6985--6996, 2021.

\bibitem{NIPS2014_5ca3e9b1}
Ian Goodfellow, Jean Pouget-Abadie, Mehdi Mirza, Bing Xu, David Warde-Farley,
  Sherjil Ozair, Aaron Courville, and Yoshua Bengio.
\newblock Generative adversarial nets.
\newblock In {\em Advances in Neural Information Processing Systems (NeurIPS)},
  pages 2672--2680, 2014.

\bibitem{NIPS2016_5b69b9cb}
Di He, Yingce Xia, Tao Qin, Liwei Wang, Nenghai Yu, Tie-Yan Liu, and Wei-Ying
  Ma.
\newblock Dual learning for machine translation.
\newblock In {\em Advances in Neural Information Processing Systems (NeurIPS)},
  pages 820--828, 2016.

\bibitem{he2020fastreid}
Lingxiao He, Xingyu Liao, Wu Liu, Xinchen Liu, Peng Cheng, and Tao Mei.
\newblock Fastreid: A pytorch toolbox for general instance re-identification.
\newblock {\em arXiv preprint arXiv:2006.02631}, 2020.

\bibitem{FID}
Martin Heusel, Hubert Ramsauer, Thomas Unterthiner, Bernhard Nessler, and Sepp
  Hochreiter.
\newblock Gans trained by a two time-scale update rule converge to a local nash
  equilibrium.
\newblock In {\em Advances in Neural Information Processing Systems (NeurIPS)},
  pages 6629--6640, 2017.

\bibitem{hudson2021ganformer}
Drew~A Hudson and C.~Lawrence Zitnick.
\newblock Generative adversarial transformers.
\newblock {\em International Conference on Machine Learning (ICML)}, 2021.

\bibitem{NIPS2015_33ceb07b}
Max Jaderberg, Karen Simonyan, Andrew Zisserman, and koray kavukcuoglu.
\newblock Spatial transformer networks.
\newblock In {\em Advances in Neural Information Processing Systems (NeurIPS)},
  pages 2017--2025, 2015.

\bibitem{jiang2021transgan}
Yifan Jiang, Shiyu Chang, and Zhangyang Wang.
\newblock Transgan: Two transformers can make one strong gan.
\newblock {\em arXiv preprint arXiv:2102.07074}, 2021.

\bibitem{feifei}
Justin Johnson, Alexandre Alahi, and Li Fei-Fei.
\newblock Perceptual losses for real-time style transfer and super-resolution.
\newblock In {\em European Conference on Computer Vision (ECCV)}, pages
  694--711, 2016.

\bibitem{adam}
Diederik~P. Kingma and Jimmy Ba.
\newblock Adam: {A} method for stochastic optimization.
\newblock In {\em International Conference on Learning Representations (ICLR)},
  2015.

\bibitem{VAE}
Diederik~P. Kingma and Max Welling.
\newblock Auto-encoding variational bayes.
\newblock In {\em International Conference on Learning Representations (ICLR)},
  2014.

\bibitem{DIAF}
Yining Li, Chen Huang, and Chen~Change Loy.
\newblock Dense intrinsic appearance flow for human pose transfer.
\newblock In {\em Proceedings of the IEEE Conference on computer Vision and
  Pattern Recognition (CVPR)}, pages 3688--3697, June 2019.

\bibitem{9470916}
Wenqi Liang, Guangcong Wang, Jianhuang Lai, and Xiaohua Xie.
\newblock Homogeneous-to-heterogeneous: Unsupervised learning for rgb-infrared
  person re-identification.
\newblock {\em IEEE Transactions on Image Processing}, 30:6392--6407, 2021.

\bibitem{DeepFashion}
Ziwei Liu, Ping Luo, Shi Qiu, Xiaogang Wang, and Xiaoou Tang.
\newblock Deepfashion: Powering robust clothes recognition and retrieval with
  rich annotations.
\newblock In {\em Proceedings of the IEEE Conference on computer Vision and
  Pattern Recognition (CVPR)}, pages 1096--1104, June 2016.

\bibitem{SPIG}
Zhengyao Lv, Xiaoming Li, Xin Li, Fu Li, Tianwei Lin, Dongliang He, and
  Wangmeng Zuo.
\newblock Learning semantic person image generation by region-adaptive
  normalization.
\newblock In {\em Proceedings of the IEEE Conference on computer Vision and
  Pattern Recognition (CVPR)}, pages 10806--10815, June 2021.

\bibitem{PG2}
Liqian Ma, Xu Jia, Qianru Sun, Bernt Schiele, Tinne Tuytelaars, and Luc
  Van~Gool.
\newblock Pose guided person image generation.
\newblock In {\em Advances in Neural Information Processing Systems (NeurIPS)},
  pages 405--415, 2017.

\bibitem{miyato2018spectral}
Takeru Miyato, Toshiki Kataoka, Masanori Koyama, and Yuichi Yoshida.
\newblock Spectral normalization for generative adversarial networks.
\newblock In {\em International Conference on Learning Representations (ICLR)},
  2018.

\bibitem{DIST}
Yurui Ren, Xiaoming Yu, Junming Chen, Thomas~H. Li, and Ge Li.
\newblock Deep image spatial transformation for person image generation.
\newblock In {\em Proceedings of the IEEE Conference on computer Vision and
  Pattern Recognition (CVPR)}, pages 7687--7696, 2020.

\bibitem{Unet}
Olaf Ronneberger, Philipp Fischer, and Thomas Brox.
\newblock U-net: Convolutional networks for biomedical image segmentation.
\newblock In {\em Medical Image Computing and Computer-Assisted Intervention
  (MICCAI)}, pages 234--241, 2015.

\bibitem{ruder2017overview}
Sebastian Ruder.
\newblock An overview of multi-task learning in deep neural networks.
\newblock {\em arXiv preprint arXiv:1706.05098}, 2017.

\bibitem{DSC}
Aliaksandr Siarohin, Enver Sangineto, Stéphane Lathuilière, and Nicu Sebe.
\newblock Deformable gans for pose-based human image generation.
\newblock In {\em Proceedings of the IEEE Conference on computer Vision and
  Pattern Recognition (CVPR)}, pages 3408--3416, June 2018.

\bibitem{VGG}
Karen Simonyan and Andrew Zisserman.
\newblock Very deep convolutional networks for large-scale image recognition.
\newblock In {\em International Conference on Learning Representations (ICLR)},
  2015.

\bibitem{tabejamaat2021guided}
Mohsen Tabejamaat, Farhood Negin, and Francois Bremond.
\newblock Guided flow field estimation by generating independent patches.
\newblock {\em British Machine Vision Conference (BMVC)}, 2021.

\bibitem{XingGAN}
Hao Tang, Song Bai, Li Zhang, Philip H.~S. Torr, and Nicu Sebe.
\newblock Xinggan for person image generation.
\newblock In {\em European Conference on Computer Vision (ECCV)}, pages
  717--734, 2020.

\bibitem{touvron2020deit}
Hugo Touvron, Matthieu Cord, Matthijs Douze, Francisco Massa, Alexandre
  Sablayrolles, and Herv\'e J\'egou.
\newblock Training data-efficient image transformers \& distillation through
  attention.
\newblock {\em arXiv preprint arXiv:2012.12877}, 2020.

\bibitem{ulyanov2017improved}
Dmitry Ulyanov, Andrea Vedaldi, and Victor Lempitsky.
\newblock Improved texture networks: Maximizing quality and diversity in
  feed-forward stylization and texture synthesis.
\newblock In {\em Proceedings of the IEEE Conference on computer Vision and
  Pattern Recognition (CVPR)}, pages 4105--4113, July 2017.

\bibitem{transformer}
Ashish Vaswani, Noam Shazeer, Niki Parmar, Jakob Uszkoreit, Llion Jones,
  Aidan~N Gomez, \L~ukasz Kaiser, and Illia Polosukhin.
\newblock Attention is all you need.
\newblock In {\em Advances in Neural Information Processing Systems (NeurIPS)},
  pages 6000--6010, 2017.

\bibitem{Wang_Lai_Huang_Xie_2019}
Guangcong Wang, Jianhuang Lai, Peigen Huang, and Xiaohua Xie.
\newblock Spatial-temporal person re-identification.
\newblock {\em Proceedings of the AAAI Conference on Artificial Intelligence},
  33(01):8933--8940, 2019.

\bibitem{8025424}
Guangcong Wang, Jianhuang Lai, and Xiaohua Xie.
\newblock P2snet: Can an image match a video for person re-identification in an
  end-to-end way?
\newblock {\em IEEE Transactions on Circuits and Systems for Video Technology},
  28(10):2777--2787, 2018.

\bibitem{ef88fbee6d014487a9995842fd8a6f64}
Yiren Wang, Yingce Xia, Tianyu He, Fei Tian, Tao Qin, {Cheng Xiang} Zhai, and
  {Tie Yan} Liu.
\newblock Multi-agent dual learning.
\newblock In {\em International Conference on Learning Representations (ICLR)},
  2019.

\bibitem{DTL}
Yijun Wang, Yingce Xia, Li Zhao, Jiang Bian, Tao Qin, Guiquan Liu, and Tie-Yan
  Liu.
\newblock Dual transfer learning for neural machine translation with marginal
  distribution regularization.
\newblock In {\em Proceedings of the AAAI Conference on Artificial
  Intelligence}, pages 5553--5560, 2018.

\bibitem{SSIM}
Zhou Wang, A.C. Bovik, H.R. Sheikh, and E.P. Simoncelli.
\newblock Image quality assessment: from error visibility to structural
  similarity.
\newblock {\em IEEE Transactions on Image Processing}, 13(4):600--612, 2004.

\bibitem{PISE}
Jinsong Zhang, Kun Li, Yu-Kun Lai, and Jingyu Yang.
\newblock Pise: Person image synthesis and editing with decoupled gan.
\newblock In {\em Proceedings of the IEEE Conference on computer Vision and
  Pattern Recognition (CVPR)}, pages 7982--7990, June 2021.

\bibitem{9623376}
Quan Zhang, Jianhuang Lai, Zhanxiang Feng, and Xiaohua Xie.
\newblock Seeing like a human: Asynchronous learning with dynamic progressive
  refinement for person re-identification.
\newblock {\em IEEE Transactions on Image Processing}, 31:352--365, 2022.

\bibitem{9540780}
Quan Zhang, Jianhuang Lai, and Xiaohua Xie.
\newblock Learning modal-invariant angular metric by cyclic projection network
  for vis-nir person re-identification.
\newblock {\em IEEE Transactions on Image Processing}, 30:8019--8033, 2021.

\bibitem{Zhang_2018_CVPR}
Richard Zhang, Phillip Isola, Alexei~A. Efros, Eli Shechtman, and Oliver Wang.
\newblock The unreasonable effectiveness of deep features as a perceptual
  metric.
\newblock In {\em Proceedings of the IEEE Conference on computer Vision and
  Pattern Recognition (CVPR)}, pages 586--595, June 2018.

\bibitem{market1501}
Liang Zheng, Liyue Shen, Lu Tian, Shengjin Wang, Jingdong Wang, and Qi Tian.
\newblock Scalable person re-identification: A benchmark.
\newblock In {\em Proceedings of the IEEE International Conference on Computer
  Vision (ICCV)}, pages 1116--1124, 2015.

\bibitem{deformdetr}
Xizhou Zhu, Weijie Su, Lewei Lu, Bin Li, Xiaogang Wang, and Jifeng Dai.
\newblock Deformable {DETR:} deformable transformers for end-to-end object
  detection.
\newblock In {\em International Conference on Learning Representations (ICLR)},
  2021.

\bibitem{PATN}
Zhen Zhu, Tengteng Huang, Baoguang Shi, Miao Yu, Bofei Wang, and Xiang Bai.
\newblock Progressive pose attention transfer for person image generation.
\newblock In {\em Proceedings of the IEEE Conference on computer Vision and
  Pattern Recognition (CVPR)}, pages 2342--2351, 2019.

\end{thebibliography}
}

\end{document}